\documentclass{svproc}
\usepackage[utf8]{inputenc} 
\usepackage[T1]{fontenc}    
\usepackage{hyperref}       
\usepackage{url}            
\usepackage{booktabs}       
\usepackage{amsfonts}       
\usepackage{nicefrac}       
\usepackage{microtype}      

\usepackage{amsmath}
\usepackage{amssymb}
\usepackage{bm}
\usepackage{lipsum}
\usepackage{pifont}
\usepackage{graphicx}
\usepackage{subcaption}
\usepackage{caption}
\usepackage{varwidth}
\usepackage{centernot}
\usepackage{todonotes}

\usepackage{algorithm}
\usepackage{algpseudocode}

\usepackage{multicol}
\usepackage{multicol}
\usepackage{lipsum}
\usepackage[font=small,labelfont=bf]{caption}
\usepackage[normalem]{ulem} 
\usepackage{float}

\usepackage{url}

\begin{document}
\mainmatter              
\title{Amenable Sparse Network Investigator}
\titlerunning{ASNI}  
%
\author{Saeed Damadi \and Erfan nouri \and Hamed Pirsiavash }
\authorrunning{Saeed Damadi, Erfan nouri, and Hamed Pirsiavash} 
%
\tocauthor{Saeed Damadi, Erfan nouri, and Hamed Pirsiavash}
\institute{Department of Computer Science and Electrical Engineering\\
University of Maryland, Baltimore County\\
Baltimore, MD 21250\\
\email{sdamadi1,erfan1,hpirsiavash@umbc.edu}}

\maketitle              

\begin{abstract}
We present “Amenable Sparse Network Investigator” (ASNI) algorithm that utilizes a novel pruning strategy based on a sigmoid function that induces sparsity level globally over the course of one single round of training. The ASNI algorithm fulfills both tasks that current state-of-the-art strategies can only do one of them. The ASNI algorithm has two subalgorithms: 1) ASNI-I, 2) ASNI-II. ASNI-I
learns an accurate sparse off-the-shelf network only in one single round of training. ASNI-II learns a sparse network and an initialization that is quantized, compressed, and from which the sparse network is trainable.  The learned initialization is quantized since only two numbers are learned for initialization of nonzero parameters in each layer L. Thus, quantization levels for the initialization of the entire network is 2L. Also, the learned initialization is compressed because it is a set consisting of 2L numbers.
The special sparse network that can be trained from such a quantized and compressed initialization is called amenable.
For example, in order to initialize more than 25 million parameters of an amenable ResNet-50, only 2x54 numbers are needed. To the best of our knowledge, there is no other algorithm that can learn a quantized and compressed initialization from which the network is still trainable and is able to solve both pruning tasks. Our numerical experiments show that there is a quantized and compressed initialization from which the learned sparse network can be trained and reach to an accuracy on a par with the dense version. This is one step ahead towards learning an ideal network that is sparse and quantized in a very few levels of quantization.  
We experimentally show that these 2L levels of quantization are concentration points of parameters in each layer of the learned sparse network by ASNI-I. In other words, we show experimentally that for each layer of a deep neural network (DNN) there are two distinct normal-like distributions whose means can be used for initialization of an amenable network. To corroborate the above, we have performed a series of experiments utilizing networks such as ResNets, VGG-style, small convolutional, and fully connected ones on ImageNet, CIFAR10, and MNIST datasets.
\end{abstract}
\begin{figure}[t]
\centering
\includegraphics[scale=0.25]{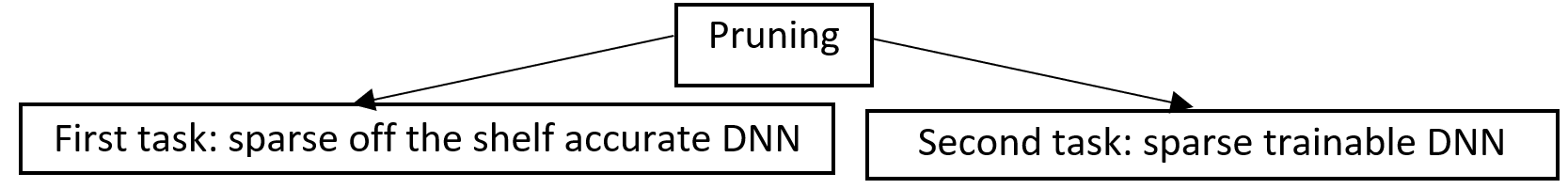}
\caption{Pruning tasks.}
\label{fig:tasks}
\end{figure}

\section{Introduction}\label{sec:intro}
As indicated in Fig. \ref{fig:tasks}, pruning of a DNN is done to fulfill two tasks: (1) obtaining an accurate sparse network, ready to use, whose test accuracy is on a par with the dense version of the network, and (2) finding trainable sparse structures that can be trained in isolation and reach test accuracy of the dense network.
The goal of the first task is to provide an accurate off-the-shelf-network that can be used in small devices, such as smart phones.
Although algorithms that solve the first task are very mature by now, solving the second pruning task is an ongoing research topic. Solving the latter task is of importance because over-parameterization \cite{denil2013predicting} has been an obstacle for interpretability of parameters in the network. Therefore, finding a sparse network that can fit to a data from scratch would help to interpret parameters of a DNN.

\begin{figure}
\centering
\includegraphics[scale=0.42]{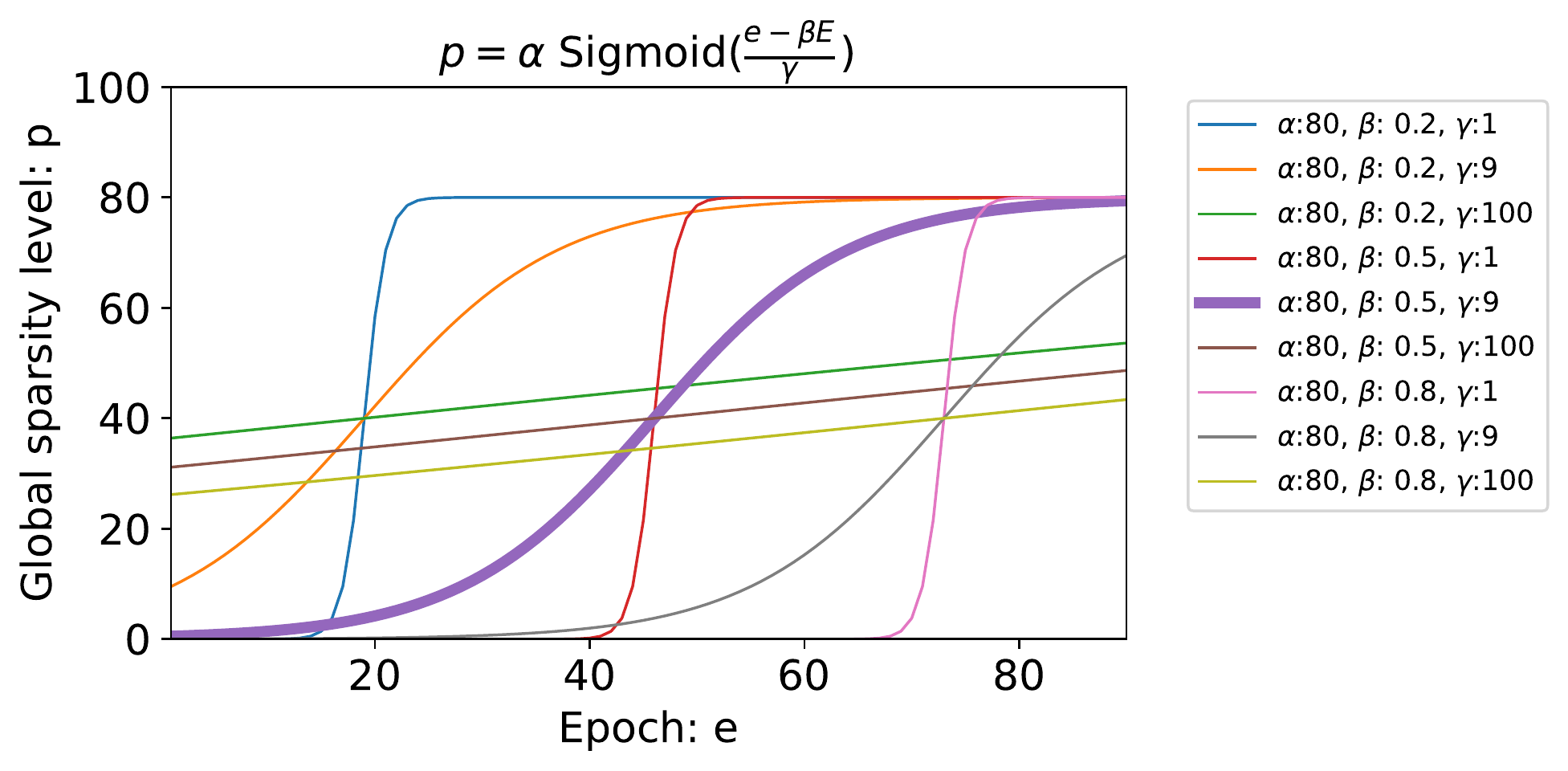}
    \caption{This figure shows different global sparsity percentages obtained from sigmoid function. The sparsity percentage is applied across the network during $E=90$ training epochs. The value of $\gamma$ determines the transition slope from a low sparsity percentage to a high one. For large $\gamma$, e.g., 100, the slope is small and almost constant during all epochs. In this case, the initial sparsity percentage is high which means a lot of pruning takes place at the beginning. For small $\gamma$, e.g., 1, the transition slope is very sharp which means at some epoch of training a lot of pruning happens.
    The best one with $\beta=0.5$ and $\gamma=E/10=9$ allows the network to learn with the full capacity at the early epochs, i.e., up to $e=10$. Then, the network sparsity starts increasing with a very small slope to make sure learning of the sparse structure is done properly, i.e., $e=10-30$. After learning the structure of the network, pruning increases with a constant slope, i.e., $e=30-60$. Then, pruning decreases very rapidly  to allow the network heal from pruning, i.e., $e=60-80$. Finally, to fine tune the survived parameters pruning is almost zero from $e=80-90$.}
    \label{fig:sigmoid}
\end{figure}

To find a trainable sparse network, \cite{frankle2018lottery} was the first work that proposes an algorithm that can solve the second task of pruning, i.e., the \textit{lottery ticket algorithm}. Since then search for finding a sparse trainable network has attracted a lot of attention.
However, the \textit{lottery ticket algorithm} (LTA) requires multiple rounds of training and it fails to find the ticket for large networks, i.e., ResNet-50.
Foresight pruning or pruning before training \cite{lee2018snip,wang2020picking,tanaka2020pruning} tries to address the first issue. However, none of foresight pruning methods perform as good as methods that obtain sparse structures using gradual pruning \cite{gale2019state}.
The stabilized LTA, \cite{frankle2019stabilizing}, shows that the second issue of LTA can be solved. By changing the initialization that uses values different than the original initialization, \cite{frankle2019stabilizing} shows a large sparse structure obtained from the stabilized LTA reaches test accuracy approximately.
The stabilized LTA uses values of the learned parameters at $k$-th iteration of the first training round. This change is equivalent to using another sparse initialization for training of a sparse network. Also, it shows that there might be some other initialization from which a sparse network is trainable. Inspired by that, we learn another initialization that is quantized, compressed, and from that the sparse network is trainable.
To this end, we present the ``Amenable Sparse Network Investigator'' (\textit{ASNI}) which performs as well as the state-of-the-art algorithms in solving the first task of pruning and solves the second task of pruning by learning an \textit{amenable} sparse network that is trainable from a learned quantized and compressed initialization. 
To the best of our knowledge there is no other algorithm that can solve both two tacks together. Also, the \textit{ASNI} algorithm is the first proposed algorithm that learns a quantized and compressed initialization from which a sparse network is trainable.
\begin{figure*}[b]
    \centering
    \includegraphics[scale=0.3]{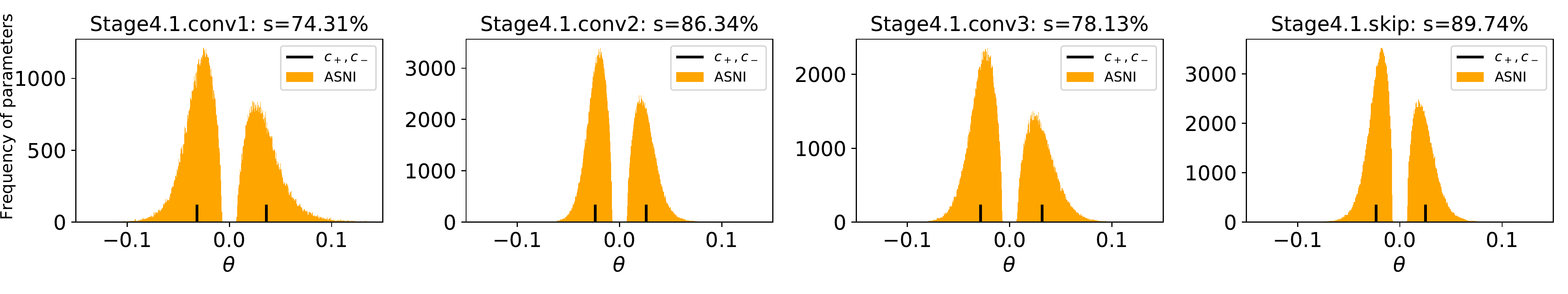}
    \caption{Parameter distribution of ResNet-50 trained on ImageNet with \textit{ASNI-I}.
    ResNet-50 has 4 main stages where stages have 3, 4, 6, and 3 bottlenecks, respectively. The first bottleneck of each stage has 3 convolution layers and a skip connection. This picture shows parameter distribution of the first bottleneck in the forth stage at overall sparsity percentage of $s=80\%$. Two short bars indicate $c_+$ and $c_-$ of each layer. These bars are the averages of parameters and used for the initialization of the learned amenable sparse network. They are found by \textit{ASNI-II} in lines 13 and 15 in Alg. \ref{alg:asni}.
    To see all stages of the network, check Appendix.
    }
    \label{fig:layer_distribution}
\end{figure*}
%
%
In summary, the following are our contributions:
\begin{itemize}
\item
We present the \textit{ASNI} algorithm (Alg. \ref{alg:asni}) that solves both pruning tasks via two subalgorithms. In only one single round of training, \textit{ASNI-I} utilizes a novel simple pruning strategy based on a sigmoid function that induces sparsity globally across the network and learns an accurate sparse network.
Fig. \ref{fig:sigmoid} shows how one can select the parameters of the sigmoid function properly to avoid harmful consequences of pruning during the training process.
\item
\textit{ASNI-II} solves the second task of pruning. It takes the output of \textit{ASNI-I} and learns an \textit{amenable} sparse network with its corresponding quantized and compressed initialization. 
We experimentally show that the learned initialization is a set of averages;
the average is taken over positive and negative learned parameters by \textit{ASNI-I} for each layer, i.e., lines 12 and 14 in Alg. \ref{alg:asni}.
Fig. \ref{fig:layer_distribution} shows the parameter distribution of the output of \textit{ASNI-I} for ResNet-50 where each layer has two distinct normal-like distributions whose means are used as the learned initialization. This pattern repeats for all networks that we study.
\item
Finally, as given in Tab. \ref{tab:prune}, we show that the \textit{amenable} sparse network equipped with the learned quantized and compressed initialization approximately achieves the test accuracy of the dense network. 
\end{itemize}
We compare \textit{ASNI-I} against its counterparts \cite{zhu2017prune,kusupati2020soft,evci2020rigging} where the two last ones are the state-of-the-art methods. We show numerically that \textit{ASNI-I}
solves the first task of pruning with higher accuracy. Also, we show that the test accuracy of a sparse \textit{amenable} network learned by \textit{ASNI-I} and initialized by the quantized and compressed initialization learned by \textit{ASNI-II} is higher than its counterparts. This is the case either with methods that solve the second task of pruning in one round \cite{frankle2020linear,frankle2020pruning} or the ones that use foresight pruning, i.e.,  \cite{lee2018snip,wang2020picking}. Hence, the \textit{ASNI} algorithm is capable of solving both tasks of pruning. 
%

\section{Related Work}\label{sec:work}
In this section we will go over related work trying to solve the first and second tasks of pruning as indicated in Fig. \ref{fig:tasks}. We focus on the second task since it is a newer task compare to the first one.
\subsection{First task of pruning}
Working on the first task of pruning started more than 30 years ago \cite{hanson1989comparing,lecun1990optimal,hassibi1993second}. However, for deep neural networks, initially \cite{han2015learning} showed that it is possible to achieve off-the shelf-sparse networks simply by magnitude-wise pruning.
Since then, many works have been done using this concept,  \cite{han2015deep,wen2016learning,guo2016dynamic,molchanov2016pruning,li2016pruning,narang2017exploring} and \cite{luo2017thinet,louizos2017learning,liu2018rethinking,tartaglione2018learning,mocanu2018scalable,dettmers2019sparse}.
The current approach for solving this task is learning an accurate sparse network in one single round of training. \cite{zhu2017prune} was the first work that tried to learn an accurate sparse network in this way. The 
current state-of-the-art \cite{kusupati2020soft} and \cite{evci2020rigging} solve this task via one single round of training. The former starts from a dense network but the latter starts from a sparse network that has desired sparsity and finds a an accurate sparse network.
\subsection{Second task of pruning}
Solving the second task of pruning is to find a sparse trainable network. To solve it, one needs to find a mask. This mask may be found in multiple rounds of training or without training. We will go over both cases.
\subsubsection{Multiple rounds of training}
To solve the second task \cite{frankle2018lottery} proposes 
a practical way of finding a sparse trainable network. The so-called \textit{lottery ticket algorithm} learns a mask in multiple rounds of training and then shows that the associated sparse network is trainable from a specific sparse initialization. This initialization is determined using the learned mask and the original initialization. However, LTA is not able to find tickets for large networks like ResNet-50. The problem is that the sparse network can not be trained from the sparse initialization.
Hence, the stabilized LTA, \cite{frankle2019stabilizing}, uses parameters of the $k$-th step from the first training round as the initialization to reach test accuracy of the dense network approximately.
Rewinding to an intermediate step $k$ in order to obtain a sparse trainable network has also been observed by \cite{you2019drawing}.
Necessity for rewinding to the $k$-th step utilized by \cite{frankle2019stabilizing} and \cite{you2019drawing} corroborates the observation in \cite{achille2018critical} that states ``in the early stages of training, important connectivity patterns among parameters of layers are discovered''. Also, it has been observed that the learned connectivity patterns stay relatively fixed in the later steps of training.

The LTA has an uncomplicated strategy to force more sparsity into the network. It applies a specific sparsity percentage at the end of each round of training. Although it is very easy to implement, it may not be an optimal strategy for reducing the number of parameters. The pruning strategy of the LTA can be improved as \cite{savarese2019winning} uses a continuous strategy to remove parameters at each round of training.
However, this is done at the expense of doubling the number of parameters.

\subsubsection{Foresight pruning: zero round of training}
\cite{frankle2018lottery}, \cite{frankle2019stabilizing}, and \cite{savarese2019winning} all search for a sparse structure utilizing multiple rounds of training which is very costly.
The best way could be finding a sparse structure even before training.
This is called foresight pruning or pruning before training.
To this end, \cite{lee2018snip} (SNIP) was the first work that posited the idea of foresight pruning which determines a mask before training.
In this approach, a \textit{connection sensitivity} score is calculated before training and parameters are removed based on the score vector.
Others have tried to find different scoring vectors.
For example, \cite{wang2020picking} introduces GraSP which utilizes Hessian-gradient product to find a score vector.
As opposed to the previous proposed methods, \cite{tanaka2020pruning} finds a sparse structure using different scoring in $n$ rounds of pruning.
\subsubsection{Multiple- and zero-round are extreme}
It is true that finding the sparse structure before training is the ideal approach and may have a negligible computational cost to find a mask, but the performance of the methods that apply pruning before training is not competitive with performance of networks whose initialization is obtained using pruning far later in training \cite{gale2019state,frankle2020pruning}.
To avoid multi-round training computation and loss accuracy in foresight pruning \cite{frankle2020pruning} uses magnitude based pruning approach that learns a sparse structure in a single round. Similarly, 
\cite{frankle2020linear} uses one round plus some $k$ steps to find the new set of initialization.
Considering these two last methods, our approach learns a sparse structure in one single round of training.

\section{Finding a sparse trainable network}\label{sec:method}
In this section we first define the exact optimization problem that solves the first task of pruning. Then, we elaborate on  how changing initialization results in solving the second task of pruning.
\subsection{Problem explanation}\label{sec:problem}
Finding a sparse network whose accuracy is on a par with a dense network amounts to solving a bi-level, constrained, stochastic, nonconvex, and non-smooth sparse optimization problem as follows:
\begin{equation}\label{eq:1}
\begin{aligned}
\bm{\theta}*
&=
\arg\min_{\hat{\bm{\theta}}}||\hat{\bm{\theta}}||_0\\
&
\text{ s.t. }
\hat{\bm{\theta}} 
=
\arg \min_{\bm{\theta}}
\mathbb{E}_{\mathbf{x}\sim \mathcal{D}}
\Big[
f
\Big(
\mathbf{y}(\mathbf{x}),
\boldsymbol{h}(\mathbf{x};\bm{\theta})
\Big)
\Big]
\end{aligned}
\end{equation} 
where $\boldsymbol{h}(\mathbf{x};\bm{\theta})$ is the vector-valued neural network function whose input is a random vector $\mathbf{x}\sim \mathcal{D}$ labeled by random vector $\mathbf{y}(\mathbf{x})$ and its parameters are denoted by $\bm{\theta}\in \mathbb{R}^d$. The scalar-valued function $f$ is the cost or loss function, also known as the criterion,  $||\bm{\theta}||_0$ is the number of nonzero elements of $\bm{\theta}$, or the $\ell_0$ norm \footnote{$\ell_0$ is not mathematically a norm because for any norm $\|\cdot \|$ and $\alpha \in \mathbb{R}$, $\|\alpha \bm{\theta} \| = \vert \alpha \vert \|\bm{\theta}\|$, while $\|\alpha \bm{\theta} \|_0 = \vert \alpha \vert \|\bm{\theta}\|_0$ if and only if $\vert \alpha \vert = 1$}. In Problem $(\ref{eq:1})$: having two optimizations simultaneously makes it bi-level, $\mathcal{D}$ is the source of stochasticity and is unknown, composition of $f$ and $\boldsymbol{h}$ make the objective function nonconvex; since $\ell_0$ norm is not differentiable, one deals with a non-smooth problem.
If one can solve Problem $(\ref{eq:1})$, the energy consumption reduces, hardware requirements are relaxed, and performing inference become faster. 
Unfortunately, even a deterministic and convex sparse optimization problem, e.g., least-squares problem, is a combinatorial and NP-hard problem \cite{davis1994adaptive,natarajan1995sparse}. Therefore, the best approach is to convert the current bi-level optimization problem into another optimization problem that is neither bi-level, stochastic, nor non-smooth. Solving the new optimization problem will find an approximate solution to the original Problem (\ref{eq:1}), i.e., $\bm{\hat{\theta}}*$. However, the unanswered question is how to convert Problem (\ref{eq:1}) to a solvable approximate problem. By trial and error, one can find an upper bound for the sparsity of the vector parameter, i.e., $||\bm{\theta}*||_0 \lessapprox \hat{s}$. 
Given the sparsity level $\hat{s}$, we get the following optimization problem:
\begin{equation}\label{eq:2}
\begin{aligned}
(\tilde{\bm{\theta}}*\odot\tilde{\bm{m}}*) 
&=
\arg \min_{\bm{\theta},\bm{m}}
\mathbb{E}_{\mathbf{x}\sim \mathcal{D}}
\Big[
f
\Big(
\mathbf{y}(\mathbf{x}),
\boldsymbol{h}(\mathbf{x};\bm{\theta}\odot\bm{m})
\Big)
\Big]
\\
\text{s.t.} 
\quad
&
||\bm{m}||_0\leq \hat{s}
\end{aligned}
\end{equation} 
where $\bm{m}=\{0,1\}^d$ is a binary mask and $\odot$ denotes Hadamard product operator. Unlike Problem (\ref{eq:1}) where a sparsity level is automatically found, i.e., $s*$, here sparsity level $\hat{s}$ is given. Because of this fact, solutions to Problem (\ref{eq:1}) and (\ref{eq:2}), that are $\bm{\theta}*$ and  $\tilde{\bm{\theta}}*\odot\tilde{\bm{m}}*$,  may not be the same.
Having an accurate estimate for $\hat{s}\gtrapprox s*$  relaxes Problem (\ref{eq:1}) from being a bi-level optimization problem to a single level problem. However, all the other issues mentioned above still stay with Problem (\ref{eq:2}). Now from the perspective of optimization, the first question would be whether this problem is feasible or not. Empirically, \cite{frankle2018lottery} addressed this problem for the first time and conjectures that such a solution exists and named their conjecture the \textit{lottery ticket hypothesis}. Mathematically speaking, the \textit{lottery ticket hypothesis} conjectures that Problem (\ref{eq:2}) for $\hat{s}\gtrapprox s*$ is feasible and there exists a solution to that.

Once we know the solution exists, the most straightforward approach to find an approximate solution to Problem (\ref{eq:2}) is to first find a mask $\hat{\bm{m}}$ which is a good estimate of $\tilde{\bm{m}}*$. To estimate $\hat{\bm{m}}$, there are two extreme approaches. One approach finds $\hat{\bm{m}}$ using multiple rounds of training \cite{frankle2018lottery,frankle2019stabilizing,savarese2019winning}, and the other finds it before training \cite{lee2018snip,wang2020picking,tanaka2020pruning}. As opposed to the latter, the former reaches test accuracy of the dense network \cite{gale2019state,frankle2020pruning}, but it is computationally expensive. Once an accurate mask, i.e., $\hat{\bm{m}} \approx \tilde{\bm{m}}*$ is at hand, the non-smoothness and constraint of Problem (\ref{eq:2}) can be relaxed. Also, by assuming large sample size and using stochastic approximation of the expected value in the objective function \cite{ghadimi2013stochastic}, one can use the following unconstrained optimization problem as a relaxation for Problem (\ref{eq:2}):
\begin{equation}\label{eq:3}
\begin{aligned}
(\hat{\bm{\theta}}*\odot\hat{\bm{m}})  
&=
\arg \min_{\bm{\theta}}
R(\mathbf{X};\bm{\theta}\odot\hat{\bm{m}})
\end{aligned}
\end{equation} 
where $R(\mathbf{X};\bm{\theta}\odot\hat{\bm{m}})
$
is the approximation of the expected value as
$$
M^{-1}
\sum_{i=1}^M
f
\Big(
\pmb{y}(\pmb{x}^{(i)}),
\boldsymbol{h}(\pmb{x}^{(i)};\bm{\theta}\odot\hat{\bm{m}})
\Big),$$
$\mathbf{X}
$
is the data matrix as
$
\begin{bmatrix}
\pmb{x}^{(1)}, \dots, \pmb{x}^{(M)} 
\end{bmatrix}$,
$\pmb{x}^{(i)}$ for $i=1,\dots,M$ is a realization of $\mathbf{x}\sim \mathcal{D}$, and $\pmb{y}(\pmb{x}^{(i)})$ is the realized target associated with $\pmb{x}^{(i)}$. 
\subsection{Initialization of the stabilized LTA vs LTA}\label{sssec:initialization}
%
\begin{algorithm}[t]
\caption{The LTA, \cite{frankle2018lottery}}
\label{alg:lta}
\begin{algorithmic}[1]
\Require $\bm{\theta}^0$, $\hat{\bm{m}}=\bm{1}$, $T$, $p\%$, $r$
\For{1 to r}
\State
Optimize $R(\mathbf{X};\bm{\theta}^0\odot\hat{\bm{m}})$ for T steps
\State
Keep $p$-th percentile of $|\bm{\theta}^T|\neq 0
\text{ and update } \hat{\bm{m}}
$
\State
Rewind nonzero values of $\bm{\theta}^T$ to $\bm{\theta}^0$ 
 and update $\bm{\theta}^0$
\EndFor
\end{algorithmic}
\end{algorithm}
%
\begin{algorithm}[t]
\caption{The stabilized LTA, \cite{frankle2019stabilizing}}
\label{alg:stabilizedLTA}
\begin{algorithmic}[1]
\Require $\bm{\theta}^0$, $\hat{\bm{m}}=\bm{1}$, $T$, $p\%$, $k$, $r$
\For{j=1 to r}
\State
Optimize $R(\mathbf{X};\bm{\theta}^0\odot\hat{\bm{m}})$ for T steps
\If{$j=1$}{$\quad \text{save } \tilde{\bm{\theta}}^0
\leftarrow
\bm{\theta}^k$} 
\EndIf
\State
Keep $p$-th percentile of $|\bm{\theta}^T|\neq 0
\text{ and update } \hat{\bm{m}}
$
\State
Rewind nonzero values of $\bm{\theta}^T$ to $\tilde{\bm{\theta}}^0$ 
 and create $\bm{\theta}^0$
\EndFor
\end{algorithmic}
\end{algorithm}
%
%
Given a mask $\hat{\bm{m}}$, every initialization from which Problem (\ref{eq:3}) can be solved iteratively is an acceptable initialization. The LTA algorithm in Alg. \ref{alg:lta}, proposes an initialization that is based on the original initialization. On the other hand, the stabilized LTA in Alg. \ref{alg:lta} uses the learned parameters of the $k$-th step of training as the initialization. Our algorithm will learn another acceptable initialization using \textit{ASNI-II}. 

To elaborate on the importance of the initialization notice that the LTA proposes Alg. \ref{alg:lta} that solves Problem (\ref{eq:3}) for $r$ rounds when $\hat{\bm{m}}$ is assumed to be given in each round. It starts from a dense random initialization and a mask whose elements are all one, i.e., $\hat{\bm{m}}=\bm{1}$. Then, it trains the network for $T$ steps. At the end of training $p\%$ of nonzero parameters are zerod out and mask gets updated. Finally, nonzero parameters are rewound to their associated entries in the original initialization to update the initialization for the next round of training.

The issue with Alg. \ref{alg:lta} is that it fails to find the winner ticket for large networks such as ResNet-50. This issue is  solved by the stabilized LTA given in Alg. \ref{alg:stabilizedLTA}.
As one can see in Alg. \ref{alg:lta}, LTA rewinds the learned parameters to the original initialization while the stabilized LTA rewinds the learned parameters to the $k$-th step of the first round of training. This little tweak stabilizes the LTA and makes it possible to find winning tickets for large networks. 
As we will see later, \textit{ASNI-II} rewinds the nonzero parameters to the average of parameters learned by \textit{ASNI-I}. This rewind is done so that sign of each learned initialization element is in accordance with learned parameters by \textit{ASNI-I}.
%
\section{Method}
\begin{algorithm}[t]
\caption{The ASNI}
\label{alg:asni}
\begin{algorithmic}[1]
\Require
Initial parameter vector $\boldsymbol{\theta}^0$, training data $\mathbf{X}_{tr}$, epochs $E$, optimizer's parameters, e.g., initial learning rate $\eta=\eta_0$, cosine scale $\delta$, mini-batch size $b$, initial mask vector $\hat{\bm{m}}=\bm{1}$, sigmoid's parameters $\alpha$, $\beta$, $\gamma$.
\State
\textbf{ASNI-I:}
 \For{$e = 1$ to $E$}
 \For{$k = 1$ to $T$}
 \State
 $\displaystyle{\bm{\theta}^{k} 
 \leftarrow 
 {\bm{\theta}}^{k-1}-\eta\left( \nabla R(\mathbf{X}_{tr,b};\bm{\theta}^{k-1})\right)\odot\hat{\bm{m}}}$
 \EndFor
 \State $\displaystyle{p=\alpha \, \text{sigmoid}((e-\beta E)/\gamma)}$
\State
 $\displaystyle{\tau_g = p\text{-th} \text{ percentile of } \{|\bm{\theta}|\}}$
 \State
 $\displaystyle{\hat{\bm{m}}=\bm{1}_{|s|\geq \tau_g}
 (|\mathbf{s}| )}$ 
\State
 $\displaystyle{\bm{\theta}^0 \leftarrow \bm{\theta}^T \odot \hat{\bm{m}}}$
 \EndFor
%
 \State
 $\displaystyle{\hat{\bm{\theta}}* \leftarrow \bm{\theta}^0}$
\State
\textbf{ASNI-II:}
 \For {$l = 1$ to $L$}
 \State
 $\displaystyle{\bar{c}_+^{[l]}=\text{mean}(
 {\hat{\bm{\theta}}*}^{[l]}>0)}$
 \State
 $\displaystyle{{\bm{\theta}^{[l]}}_+^0=\bar{c}_+^{[l]}
 \bm{1}_{
 {\hat{\bm{\theta}}*}^{[l]}>0
 }
 ({\hat{\bm{\theta}}*}^{[l]})
 }
 $
 \State
 $\displaystyle{\bar{c}_-^{[l]}=\text{mean}(
 {\hat{\bm{\theta}}*}^{[l]}<0)}$
 \State
 $\displaystyle{{\bm{\theta}^{[l]}}_{-}^0=\bar{c}_{-}^{[l]}
 \bm{1}_{
 {\hat{\bm{\theta}}*}^{[l]}<0
 }
 ({\hat{\bm{\theta}}*}^{[l]})
 }
 $
 \State
 $\displaystyle{{\bm{\theta}^{[l]}}^0={\bm{\theta}^{[l]}}_{-}^0
 +
 {\bm{\theta}^{[l]}}_{+}^0}
 $
 \EndFor
 \State
$\displaystyle{\textbf{Output:\quad} \hat{\bm{{\theta}}}*, \bm{{\theta}}^0, \bar{\mathbf{c}}_+, \bar{\mathbf{c}}_- \text{~and~} \hat{\bm{m}}}$
\end{algorithmic}
\end{algorithm}
In this section we introduce the \textit{ASNI} algorithm and explain how one can set its parameters.
\subsection{The ASNI algorithm}\label{sssec:asni}
The \textit{ASNI} algorithm first learns an off-the-shelf accurate sparse structure via its subalgorithms \textit{ASNI-I}.
This learned sparse structure is trainable. Its trainability depends on $L$ pairs of signed centroids that are sufficient for its initialization. Being able to be trained from a compressed set of initialization, makes the learned sparse network \textit{amenable}. 
As Fig.  (\ref{fig:layer_distribution}) shows, after learning parameters by \textit{ASNI-I}, distribution of each layer $l$ has a pair of signed centroids (positive and negative), i.e., $\bar{\mathbf{c}}_+^{[l]}$, $\bar{\mathbf{c}}_{-}^{[l]}$. These centroids are at the mean values of two distinct normal-like distributions. Once these centroids are replaced instead of positive and negative learned values, the sparse network is \textit{amenable} and reaches virtually the nominal test accuracy of the dense network. Hence, the second task of pruning is solved using \textit{ASNI-II}.

One crucial part of the \textit{ASNI} algorithm is where the sigmoid function is utilized because it determines the global sparsity percentage. This sparsity percentage is initialized to be zero and is updated at line 5 of Alg. (\ref{alg:asni}). According to line 6, \textit{ASNI-I} prunes nonzero parameters of the network with respect to a global pruning threshold, i.e., $\tau_g > 0$.  This global pruning threshold is obtained after each epoch of training. The global threshold is calculated by gathering magnitudes of all parameters except the bias and batch normalization parameters. We do not include bias and batch normalization parameters since the number of these parameters is negligible. The global threshold, i.e., $\tau_g$, is found such that the magnitude of $p\%$ of all gathered parameters are less than $\tau_g$, and $(100-p)\%$ are above the global threshold. Next in line 7 \footnote{$\bm{1}_{A}(x)=\{ 1\text{ if } x\in A \text{,} 0 \text{ if } x\notin A\}$}, \textit{ASNI-I} zeros out the entries of the mask vector for those parameters whose magnitudes are less than $\tau_g$. This is where the mask vector $\hat{\bm{m}}$ gets updated. Once \textit{ASNI-I} updates the mask, retraining restarts for another epoch. As the last epoch finishes, the vector of parameters would be the learned sparse vector, i.e., $\hat{\bm{{\theta}}}*$. This learned sparse vector together with its mask is an approximate solution to Problem (\ref{eq:2}). When $\bm{\hat{\theta}}*$ is found, by following lines 11-16, centroids ($\bar{\mathbf{c}}_+^{[l]}, \bar{\mathbf{c}}_{-}^{[l]}$) are calculated to learn the quantized and compressed initialization. Also, the mask $\hat{\bm{m}}$ learned by \textit{ASNI-I} identifies the sparse \textit{amenable} network. 
Note that initialization for batch normalization weights is one and initialization for all biases would be zero. 
\subsection{ASNI parameters}\label{sec:anatomy}
\textit{ASNI} algorithm follows a simple intuitive and easy strategy for determining the global sparsity percentage for each epoch. In line 5, \textit{ASNI-I} determines sparsity percentage by utilizing a sigmoid function as 
$p=\alpha \, \text{sigmoid}((e-\beta E)/\gamma)$
for $e=1,\dots,E$, where $E$ is the total number of epochs,  $\alpha$ controls the final sparsity, $\beta$ governs how early and late pruning starts and stops, and $\gamma$ controls how fast pruning should be done.
Although the sigmoid function has three parameters, by determining two of them the last one is determined. Thus, we only need to search for two parameters.
We will explain how to chose these parameters in the order of their importance. Hence, we start with the most important parameter which is $\beta$.
\subsubsection{How to choose $\beta$?}
To apply the \textit{ASNI-I} algorithm one needs to set $\beta$ to 0.5 for all experiments. The value of $\beta$ shifts the position of the inflection point of the sigmoid curve to the left or right. In Fig. \ref{fig:sigmoid}, the inflection point is at $45^{th}$ epoch. Therefore, up to that point, the sparsity percentage curve is increasing while after $45^{th}$ epoch the sparsity percentage decreases.
Setting $\beta$ to 0.5 creates a symmetric curve about the inflection point of sigmoid function. Generally, every sparsity curve like the ones in Fig. \ref{fig:sigmoid} has three phases: 1) small pruning associated with learning the structure, 2) pruning and learning, 3) small pruning for healing from pruning. Any value other than 0.5 will be biased towards either phase 1 or 3.
\subsubsection{How to choose $\gamma$?} 
The value of $\gamma$ determines the transition slope from a low sparsity percentage to a high one. The lower $\gamma$, the higher transition slope form a mild pruning strategy to an aggressive one. For a small $\gamma$, e.g., 1, the slope in Fig. \ref{fig:sigmoid} is very sharp and so many parameters would be pruned in a very few number of epochs. On the other hand, for a large $\gamma$, e.g., 100, the slope in Fig. \ref{fig:sigmoid} is small and the curve becomes a line starting from high sparsity percentages. This means we have a lot of pruning at the beginning. According to our experiments, $\gamma=E/10$ accompanied by $\beta=0.5$ provides the best result for all networks we experimented.
\subsubsection{How to choose $\alpha$?}
As we explained, given two parameters the third one is determined. Therefore, by determining $\beta$ and $\gamma$, the value of $\alpha$ is determined from a desired sparsity percentage.

%
\begin{table}[t]
\centering
\caption{Datasets and networks combination together with their hyper parameters.}
  \label{tab:hyper}
\scalebox{1.1}{
\begin{tabular}{@{}l *{10}{l}@{}}
\hline
\toprule
 Comb. & Dataset & Network & Params & E & B  & LR/WD & Iter.\\
\midrule
 1 & MNIST &  FC & 266,610& 50 & 60 & 1.2e-3  & 1000 \\
\midrule
  2 & MNIST &  Conv2& 3,317,450 & 20 & 60 & 2e-4  & 1000\\
\midrule
  3 & MNIST &  Conv4& 1,933,258& 25 & 60 & 3e-4  & 1000 \\
\midrule
  4 & MNIST &  Conv6& 1,802,698& 30 & 60 & 3e-4  & 1000 \\
\midrule
  5 & CIFAR-10 &  Conv2& 4,301,642 & 20 & 60 & 2e-4 & 1000\\
\midrule
  6 & CIFAR-10 &  Conv4 & 2,425,930 & 25 & 60 & 3e-4& 1000\\
\midrule
 7 & CIFAR-10 &  Conv6& 2,262,602 & 30 & 60 & 3e-4 & 1000\\
\midrule
 8 & CIFAR-10 &  VGG-11& 9,231,114& 160 & 128 & 0.05/5e-4 & 391\\
\midrule
  9 & CIFAR-10 &  VGG-13& 9,416,010& 160 & 128 & 0.05/5e-4 & 391\\
\midrule
  10 & CIFAR-10 &  VGG-16& 14,728,266& 160 & 128 & 0.05/5e-4 &  391\\
\midrule
  11 & CIFAR-10 &  ResNet-18& 11,181,642& 160 & 128 & 0.08/5e-4 & 391 \\
\midrule
  12 & ImageNet &  ResNet-50& 25,557,032& 90 & 820 & 0.35/1e-4 & 6252 \\
\midrule
\bottomrule
\hline
\end{tabular}
}
\end{table}
\section{Experiments setup}
\label{sec:setup}
This section explains the setup of our experiments for training. Tab. \ref{tab:hyper} summarizes experiments (dataset and network combinations) and their hyper parameters including number of trainable parameters, number of epochs for training (E), mini-batch size (B), the initial learning rate together with weight decay (LR/WD), and number of iterations to complete an epoch (Iter.).  Also, we will elaborate on the learning rate policy and the optimizer choice in the following subsections.

\subsection{Datasets}
Our experiments involve image classification on well-known datasets including the MNIST \cite{lecun1998mnist},  CIFAR-10 \cite{krizhevsky2009cifar}, and the ImageNet-1K \cite{russakovsky2015imagenet}. For ImageNet-1K our training pipeline uses a standard data augmentation including random flips and crops.  
\subsection{Networks}
Network architectures that we utilize include a 3-layer fully-connected network (LeNet-300-100) \cite{lecun1998gradient} named FC, convolutional neural networks (CNNs), named Conv-2, Conv-4, and Conv-6 (small CNNs with 2/4/6 convolutional layers, same as in \cite{frankle2018lottery}). We also use  ResNet-18/50 \cite{he2016deep}. Additionally, we utilize VGG-style networks \cite{simonyan2014very} named as VGG-11/13/16 with batch normalization and an average pooling after convolution layers followed by a fully connected layer. As a result of the average pooling, parameter count of VGG-style networks decreases which mitigates the parameter inefficiency with original VGG networks.
\subsection{Optimizer}
We use stochastic gradient descent \cite{robbins1951stochastic} with momentum \cite{nesterov1983method} (SGD+M), or 
Adam \cite{kingma2014adam} as our optimizers. the Adam optimizer is used for experiments involving small networks such as FC, Conv2/4/6. The SGD+M optimizer is used for the VGG-style and ResNet networks. We set the momentum coefficient to 0.9 for all experiments the SGD+M is used. 
\begin{figure}[t]
    \centering
    \includegraphics[scale=0.4]{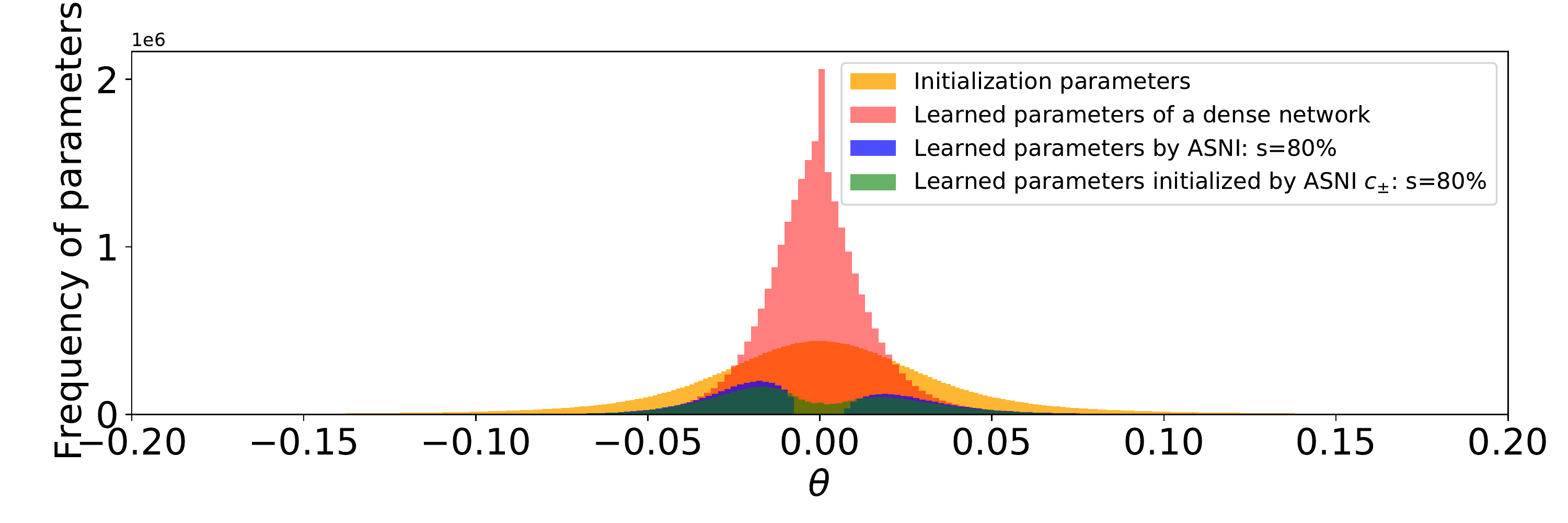}
    \caption{
  Parameter distribution considering all parameters in ResNet-50 network trained on ImageNet-1K. The orange plot shows distribution of the dense network initialization. The red plot is the distribution of the learned parameters when the network is dense. The blue one is the distribution of parameters learned by \textit{ASNI-I} at $s\approx 80\%$. The green one is the distribution of parameters for the learned sparse amenable network ($s\approx 80\%$) initialized by centroids obtained from \textit{ASNI-I}. 
    }
    \label{fig:network_distribution}
\end{figure}
\subsection{Learning rate policy}
Three learning rate policy are used for different experiments: 1) constant, 2) cosine, 3) cosine with warm-up.
\subsubsection{Constant learning rate}
For those experiments that we use the Adam optimizer, the learning rate is constant throughout the training. 
\subsubsection{Cosine policy learning rate}
For cases where the SGD+M is used, the learning rate follows a cosine policy. This cosine policy reduces the learning rate following a cosine function that has three parameters, i.e., $\eta = \eta_0\cos(\pi e/ (1+\delta)E)$ where $\eta_0$ is the initial learning rate, $E$ is the total number of epochs, and nonzero $\delta$ controls the final learning rate. Parameter $\delta$ is set to 0.05 for VGG-style and ResNet-18 networks and it is set to 0.04 for training ResNet-50. Only for training of ResNet-50 we use 10 epochs to warm up the value of learning rate. For the warm-up case learning rate increases linearly in 10 steps so that it reaches the highest value after the $10^{th}$ epoch. Then, it starts decaying following the above cosine function.
\subsection{Parameters initialization}
Network parameters are initialized according to the Kaiming Normal distribution \cite{he2015delving}.
\subsection{Deep network framework}
For all experiments we use Pytorch \cite{paszke2019pytorch} and its native automatic mixed precision (AMP) library to boost the speed of training. Networks are trained using NVIDIA TITAN X (Pascal) 12GB GPUs.
\section{Results}
\begin{table}[t]
\centering
\caption{
This table shows hyper parameters for pruning, final sparsity, and top-1 test accuracy of four variants. Accuracy percentages are the average of 5 experiments. (T1-D): dense network with zero sparsity, (T1-A-I): sparse network learned by \textit{ASNI-I}, (T1-A-II) the sparse amenable network initialized by the quantized and compressed initialization learned by \textit{ASNI-II}, (T1-S) the sparse amenable network initialized by the original initialization.
}
  \label{tab:prune}
\scalebox{1.05}{
\begin{tabular}{@{}l *{10}{l}@{}}
\hline

\toprule
Comb. &  $\alpha$ & $\gamma$ & s\%  & Nonzeros &  T1-D\% & T1-A-I\% & T1-A-II\% & T1-S\%\\
\midrule
1 & 98 & 5 & 96.87 & 8,335 & 96.88 & 96.72 &96.93 & 96.75\\
\midrule
2 & 99.2 & 2 & 98.18 & 86,363 & 98.09 & 98.12 & 98.14 & 98.03\\
\midrule
3 & 98.5 & 2 & 97.94  & 39,828 & 98.33 & 98.46 &98.52 & 98.27\\
\midrule
4 &   98.5 & 3 & 97.15 & 51,420 & 98.25 & 98.54 & 98.53 & 98.36\\
\midrule
5 &   98.5 & 2 & 96.71 & 141,364 & 76.10 & 74.61 & 76.4 & 75.26\\
\midrule
6 &  95 & 2 & 94.47 & 134,185 & 84.34 & 84 & 83.76 & 83.24\\
\midrule
7 &  94 & 3 & 92.72 & 164,624 & 86.69 & 86.1 & 85.72 & 85.55\\
\midrule
8 &  97 & 16 & 96.14 & 356,230 & 92.40 & 91.51 & 91.18 & 89.94\\
\midrule
9 &  98 & 16 & 97.14 & 269,221 & 94.23 & 92.98 & 92.29 & 91.67\\
\midrule
10 &  99 & 16 & 98.15 & 272,626 & 93.90 & 93.45 & 92.13 & 91.94 \\

\midrule
11  & 97 & 16 & 96.14 & 451,216 & 90.11 & 88.93 & 88.52 & 87.95\\

\midrule
12 &  81.04 & 10 & 80.12 & 5,080,737 & 76.08 & 75.23 & 75.17  & 70.91\\
\midrule
12 &  91.05 &  10 & 90.08 & 2,535,257 & 76.08 & 74.28 & 73.97 & 68.41\\

\bottomrule
\hline
\end{tabular}
}
\end{table}
In this section we will go over different aspects of our experiments like test accuracy, network- and layer-wise distributions of the sparse network learned by \textit{ASNI-I}, and compare \textit{ASNI-I} and learned initialization by \textit{ASNI-II} with their counterparts.

\subsection{The overall network parameters distribution}
%
%
\begin{figure*}[t]
\centering
\includegraphics[scale=0.24]{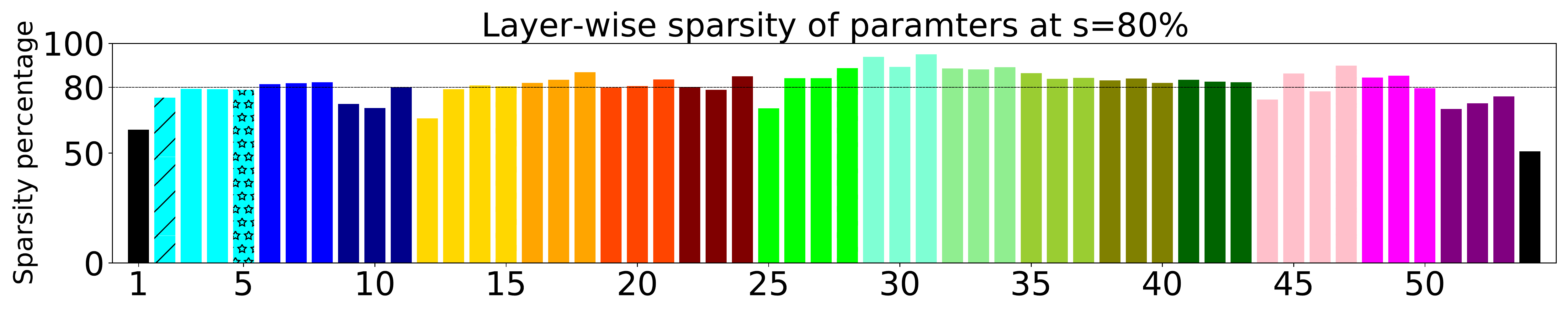}
\caption{
Layer-wise sparsity after training a ResNet-50 network. The network has 4 main stages where stages have 3, 4, 6, and 3 bottlenecks respectively. The stages are color-coded with blue (2-11), red (12-24), green (25-43), and pink (44-53). Black bars are sparsity percentage of the first convolution layer (not in stages) and the fully connected layer at the end. The first convolution layer in each stage located in the first bottleneck of that stage is hatched with lines.  Skip connections of each stage is hatched with stars. 
The least sparsity occurs at the last layer (54) which is a fully connected one. Also, first convolution layers (2, 12, 25, 44) in the first bottleneck of each stage are the ones with the smallest sparsity compared to other convolutional layers in stages. 
}
\label{fig:sparsity80}
\end{figure*}
%
The \textit{ASNI-I} algorithm uses a global threshold that considers the magnitude of every parameter across the network. Therefore, it makes sense to look at the distribution of all parameters at once and not in layer-wise fashion. Fig. \ref{fig:network_distribution} shows the network distribution for ResNet-50 trained on ImageNet-1K.  As Fig. \ref{fig:network_distribution} shows, initial parameters (orange distribution) have the largest variance. On the other hand, distribution of the learned parameters of a dense network (red distribution) has the smallest variance among all four distributions. Distribution of parameters learned by \textit{ASNI-I} (blue one) includes two normal-like distributions where small values (in the absolute value sense) have been discarded. Ignoring small values is what \textit{ASNI-I} forces but observing two normal-like distribution is surprising. The more surprising phenomenon is the parameter distribution of the \textit{amenable} sparse network initialized by \textit{ASNI-II} (green one). This distribution covers the distribution of parameters learned by \textit{ASNI-I} and fills the gap between them. This means that the learned sparse \textit{amenable} structure initialized by centroids is also able to learn the learned parameters by \textit{ASNI-I} because they have the same parameter distribution.
Compare to the layer distribution of parameters learned \textit{ASNI-I} in Fig. \ref{fig:layer_distribution}, one can observe that these normal-like distributions also exist for each layer.
That motivated us to use the averages of positive and negative values to initialize the \textit{amenable} sparse network.
\subsection{Test accuracy}
Tab. \ref{tab:prune} shows hyper parameters for pruning and test accuracy corresponding to four different variants. These four variants are: 1) dense network, a sparse network learned by \textit{ASNI-I}, the \textit{amenable} sparse network initialized by the initialization from \textit{ASNI-II}, the \textit{amenable} sparse network initialized by the original initialization.
There are two observations in Tab. \ref{tab:prune}. First, the accuracy of the dense network is not always the highest and sometimes the network learned by \textit{ASNI-I} performs better even if it has far less parameters than the dense one. Another observation is that networks initialized by the original initialization cannot reach test accuracy of the dense network. They also fall short of the accuracy of the sparse \textit{amenable} network which is initialized by the initialization learned by \textit{ASNI-II}.

\subsection{Layer-wise sparsity distribution}

\textit{ASNI-I} prunes parameters of the network to reach a predefined sparsity percentage for the entire network. After reaching  this sparsity value the question is which layers have been pruned more and which layers are denser than others. As \textit{ASNI-I} does not enforce any limitations on the sparsity of layers, each layer can be pruned differently. According to Fig. \ref{fig:sparsity80} layers are pruned non-uniformly. Among all layers the last layer (black one) is the most dense one with least pouring. The second most dense one is the very first convolutional layer.
Another interesting observation is that, the first convolutional layer in the first bottleneck of each stage is the layer with the least pruning. To observe that notice stages are color-coded with blue (2-11), red (12-24), green (25-43), and pink (44-53) in Fig. \ref{fig:sparsity80}.
\subsection{ASNI performance vs its counterparts}
The \textit{ASNI} algorithm solves both tasks of pruning in Fig. \ref{fig:tasks} simultaneously. The subalgorithm \textit{ASNI-I} solves the first task and finds an off-the-shelf accurate sparse network in one round for ResNet-50 trained on ImageNet-1k. We compare  \textit{ASNI-I} with \cite{zhu2017prune,kusupati2020soft,evci2020rigging} in Tab. \ref{tab:ASNI-I}. These methods are the ones that try to solve the first task in one round of pruning.
For the second task of the pruning we compare the accuracy of foresight pruning methods in \cite{lee2018snip,wang2020picking} and methods that utilize magnitude pruning like  \cite{frankle2020linear,frankle2020pruning} for solving the first task in one single round.


%

\section{Conclusion and discussion}
We proposed the \textit{ASNI} algorithm that solves two tasks of pruning simultaneously using two subalgorithms.
To solve the first task \textit{ASNI-I} learns an accurate sparse network in one round. This learned sparse network is \textit{amenable} because it reaches the test accuracy of the dense network starting from quantized and compressed initialization. The \textit{ASNI} algorithm owes its success to a simple pruning strategy which utilizes sigmoid function which manages the sparsity budget throughout the training process. By choosing parameters of sigmoid function properly, \textit{ASNI} symmetrically reduces pruning at the beginning and end of the pruning.

\begin{table}[t]
\centering
\caption{
First task of pruning at $s\approx 80\%$
(comb. 12)}
  \label{tab:ASNI-I}
\scalebox{1.3}{
\begin{tabular}{@{}l *{10}{l}@{}}
\hline

\toprule
Method & Nonzeros & Top-1\\
\midrule
Gradual Magnitude Pruning,  \cite{zhu2017prune}
&
5,120,000 & 74.68
\\
Soft Threshold  Reparameterization, \cite{kusupati2020soft}
&
5,120,000 & 74.87
\\
Rigging the Lottery, \cite{evci2020rigging}
&
5,120,000 & 74.55\\
\textit{ASNI-I}
&
5,080,737 & 75.23\\
\bottomrule
\hline
\end{tabular}
}
\end{table}


\begin{table}[t]
\centering
\caption{
Second task of pruning at $s\approx 80\%$
 (comb. 12)}
  \label{tab:ASNI-II}
\scalebox{1.2}{
\begin{tabular}{@{}l *{10}{l}@{}}
\hline
\toprule
Method & Nonzeros & Top-1\\
\midrule
Iterative magnitude pruning,  \cite{frankle2020linear}
&
5,120,000 & 73.33\\
Magnitude pruning after training, \cite{frankle2020pruning}
&
5,120,000 & 75.12
\\
Single-shot network pruning, \cite{lee2018snip}
&
5,120,000 & 69.29

\\
Gradient Signal Preservation, \cite{wang2020picking}
&
5,120,000 & 71.56
\\
Amenable Sparse network initialized by \textit{ASNI-II}
&
5,080,737 & 75.17\\
\bottomrule
\hline
\end{tabular}
}
\end{table}



\bibliography{neurips_2021_ansi}
\bibliographystyle{acm}

\begin{figure}
    \centering
    \includegraphics[scale=0.2]{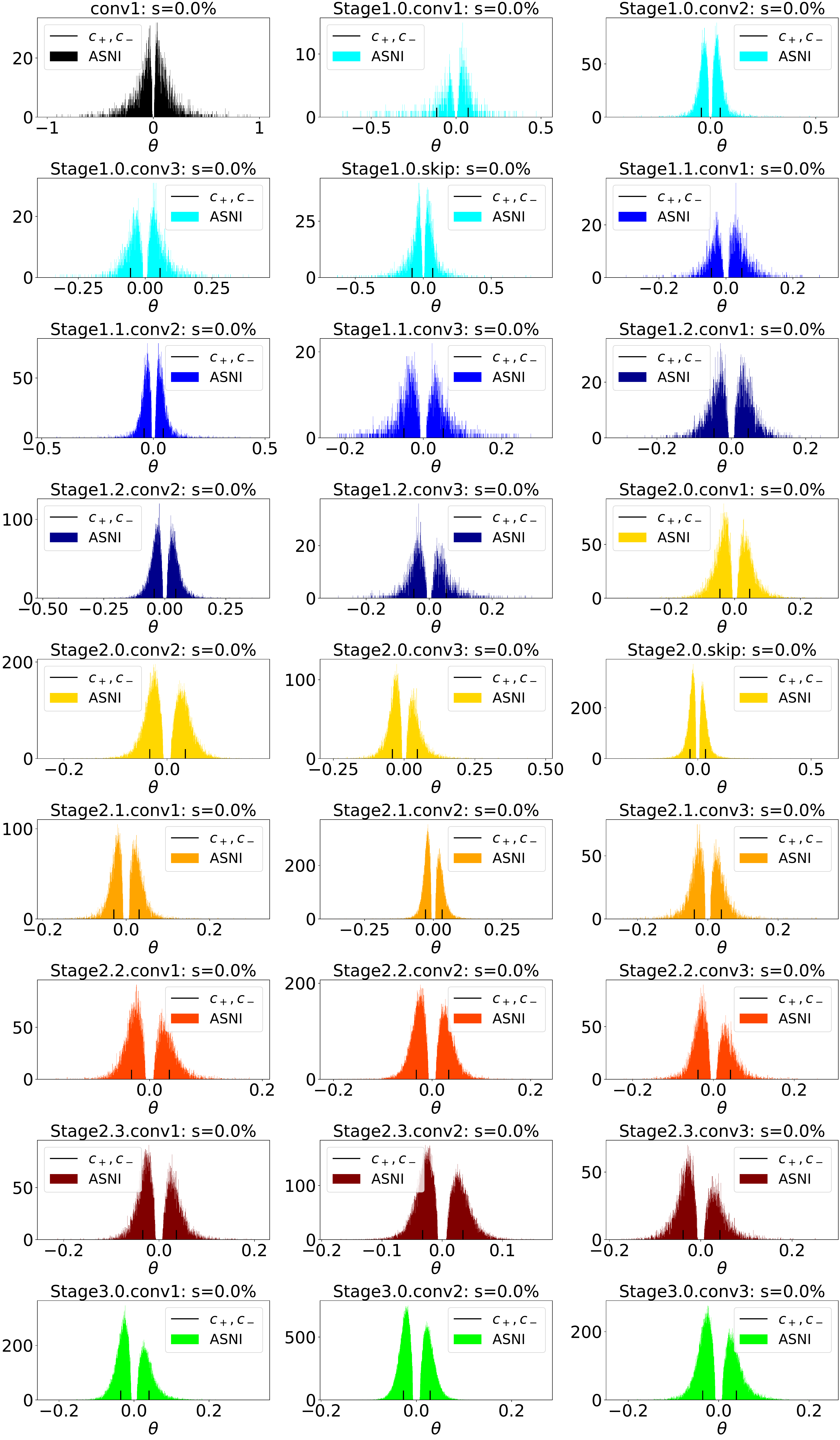}
    \caption{Appendix: ResNet50-ImageNet-1KParameter: Distribution of each layer.
    }
    \label{fig:resnet50-IMAGENET-layer1-params}
\end{figure}

\begin{figure}
    \centering
    \includegraphics[scale=0.22]{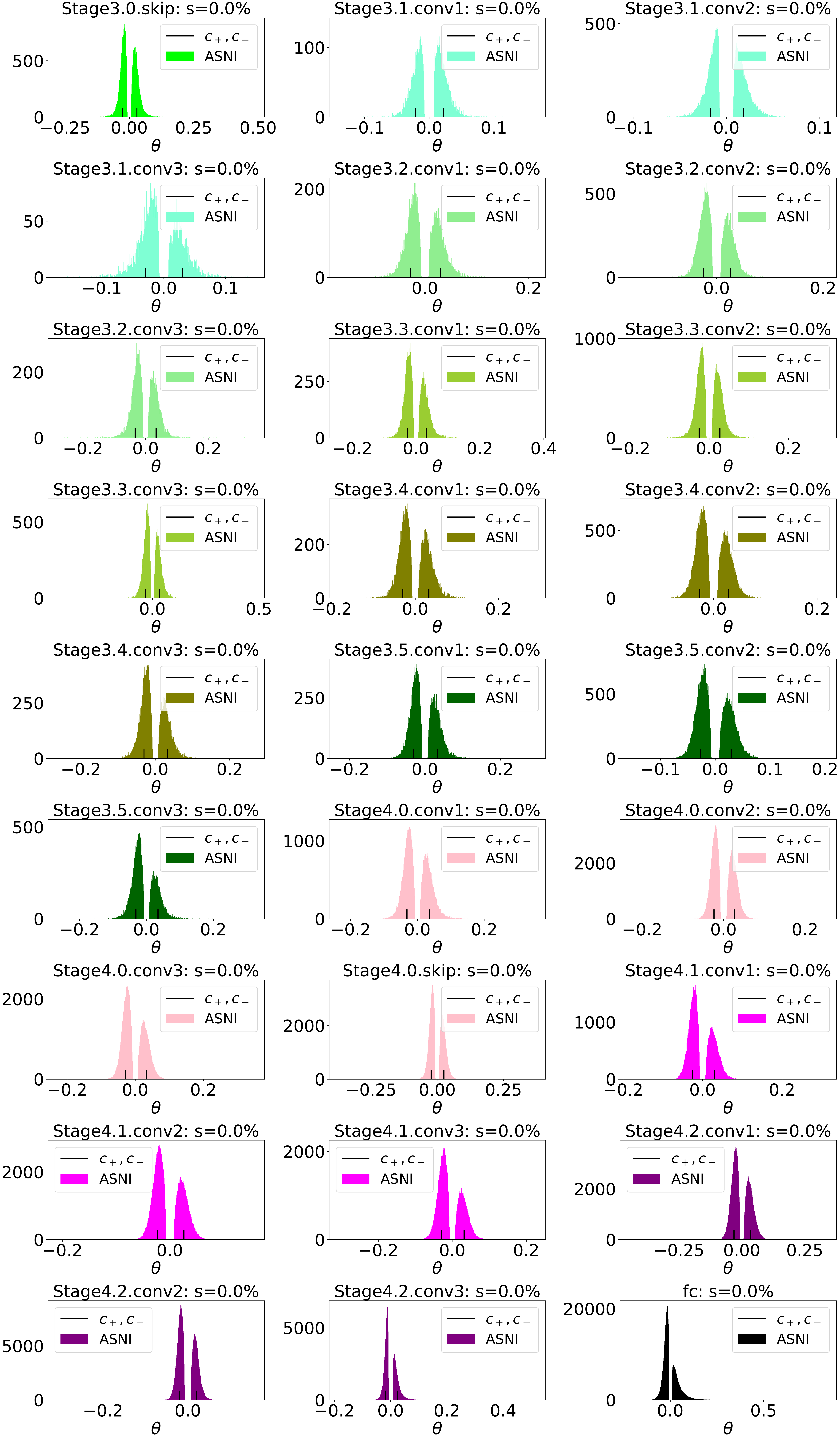}
    \caption{Appendix: ResNet50-ImageNet-1KParameter: Distribution of each layer.
    }
    \label{fig:resnet50-IMAGENET-layer2-params}
\end{figure}

\end{document}